# 3DPIFCM Novel Algorithm for Segmentation of Noisy Brain MRI Images

*Arie Agranonik, Maya Herman, Mark Last*

*Abstract*— **We present a novel algorithm named 3DPIFCM, for automatic segmentation of noisy MRI Brain images. The algorithm is an extension of a well-known IFCM (Improved Fuzzy C-Means) algorithm. It performs fuzzy segmentation and introduces a fitness function that is affected by proximity of the voxels and by the color intensity in 3D images. The 3DPIFCM algorithm uses PSO (Particle Swarm Optimization) in order to optimize the fitness function. In addition, the 3DPIFCM uses 3D features of near voxels to better adjust the noisy artifacts. In our experiments, we evaluate 3DPIFCM on T1 Brainweb dataset with noise levels ranging from 1% to 20% and on a synthetic dataset with ground truth both in 3D. The analysis of the segmentation results shows a significant improvement in the segmentation quality of up to 28% compared to two generic variants in noisy images and up to 60% when compared to the original FCM (Fuzzy C-Means).**

## I. Introduction

Image segmentation in the medical domain is a very important technique in doctor assistance systems. It's used as a first step in a multistep process in medical image analysis. The result of the segmentation has an effect on all following tasks in the pipeline. Those include feature measurement, object representation, object description and even object classification [12]. There are many methods to perform image segmentation. Those include edge detection, thresholding, region growing, clustering and supervised learning by using deep learning. Each aforementioned method has both pros and cons. No single method is considered a full solution to all modalities and image qualities. In this work we focus on clustering based approach to segmentation for several reasons. First, the methods presented in this paper are potentially applicable to different image modalities and are independent of image quality. Second, fuzzy clustering approaches are very good with medical imaging such as MRI and CT [2]. Third, we want to explore noisy medical images in 3D and see how current state of the art clustering algorithms can be improved using a 3D approach and modern optimization algorithms that can do clustering in spite of noise present in the images.

FCM [2] is one of the most popular algorithms for fuzzy clustering in the medical domain. It iterates over all pixels in the image and assigns a probabilistic score to each pixel belonging to each cluster. This way the algorithm achieves a fuzzy separation between clusters and not a hard line which is very suitable for medical images where tissues are merged. FCM performs sub optimally when there is noise in the image. This is due to its inherit nature of looking at feature attraction per pixel. Feature attraction means that each evaluation of pixels' cluster is performed by only looking at the color of all other pixels in the image and not proximate pixels. This behavior is fixed in a newer algorithm called IFCM[19] which balances between feature attraction and neighborhood attraction. The balancing is done via optimization with a neural network. Some of this algorithms' limitations are slow convergence, suboptimal results and complex neural net implementation. GAIFCM [7] solves this problem by introducing a genetic algorithm optimization with higher accuracy. It uses GA to optimize the balance between feature attraction and neighborhood attraction.

We divide our research into two stages. The first stage is to develop a new PSO[13][17] based algorithm and compare it with GA. In stage two we develop a new 3D based algorithm that utilizes surrounding voxels of the 3D image to counteract the noisy voxels in the image. We present the IFCMPSO algorithm that performs fuzzy segmentation of medical images in 2D using the particle swarm optimization [13][6]. We analyze and test the algorithm for correctness and compare it to the GAIFCM algorithm [7]. On the basis of IFCMPSO we develop the new 3DPIFCM algorithm which performs the same function as IFCMPSO but also uses the 3D voxels of each slice thus adding more information to the optimization and improving accuracy.

First, the algorithm utilizes the voxels surrounding each segmented target voxel in order to eliminate noise in the 2D slice. Second, it introduces two hyper parameters. H is used as exponential decay parameter to control how much of each surrounding voxel has an impact on the cluster depending on the distance of this voxel from the target. V is used as a depth parameter to control how many surrounding voxels to search per target voxel in 3D space during clustering. Those two parameters are evaluated in this paper to find their optimal value.

We test all algorithms on T1 Brain MRI image from Brainweb [4]. The objective of both IFCMPSO and 3DPIFCM in our test scenario is to segment the brain images into White Matter (WM), Gray Matter (GM) and Cerebral Spinal Fluid

(CSF). The IFCMPSO algorithm which works on 2D images uses a modified version of FCM for initialization of the cluster centers. The modified FCM is using a Gaussian mixture model in order to find an initial value for the cluster centers in the initialization of IFCMPSO. In our experiments we've seen that doing so avoids local minima of IFCMPSO and gives us comparable performance to GAIFCM [7].

Our contributions in this research are as follows, first we introduce a new 2D clustering algorithm. The algorithm is comparable to it's genetic variant in quality but performs less complex optimization. This is achieved by doing a smart initialization of cluster centers and using PSO. Second we introduce a new clustering algorithm in 3D that can segment 2D images in 3D space and achieves 13%-60% better accuracy results in noisy images when compared to state of the art 2D variants.

This paper is organized as follows: Section II discusses the segmentation and optimization methods IFCMPSO is based on. In Section III we introduce the proposed 3DPIFCM algorithm followed by runtime analysis in section IV. Section V contains the results of running the algorithm against Brainweb and synthetic data. Section VI contains a case study of Brain MRI. Section VII contains our conclusions and section VIII our suggestions for future work.

## II. BACKGROUND

### A. Fuzzy c-Means (FCM) and Improved Fuzzy c-Means (IFCM)

FCM is a segmentation algorithm which generalizes the c-Means algorithm, allowing soft segmentation by using fuzzy membership of each pixel to a cluster. For each pixel it assigns a membership factor to each cluster. A membership closer to 1 indicates a high degree of similarity of a pixel to other pixels in a cluster, and a value closer to 0 indicates low similarity to the data in a cluster. The FCM algorithm is an iterative clustering and produce c-partition by minimizing the weighted cost function (1) denoted as: $J_m(U, \bar{c})$

$$J_m(U, \bar{c}) = \sum_{i=1}^{N} \sum_{j=1}^{C} u_{ij}^m d^2(x_i, c_j) \qquad (1)$$

Under the following conditions:

$$0 \leq u_{ij} \leq 1 \qquad (1a)$$
$$\forall i \ \sum_{j=1}^{C} u_{ij} = 1 \qquad (1b)$$
$$\forall j \ 0 < \sum_{i=1}^{N} u_{ij} < N \qquad (1c)$$

Where:
$X \subseteq R^d$ – the data set in the $d$-dimensional vector space (for our purposes $d = 1$)

$N$ – the number of data points (pixels)
$C$ – the number of clusters

$U = (u_{ij})_{1 \leq i \leq N, 1 \leq j \leq C}$ – the membership matrix (the c-partition of $X$)

$u_{ij}$ – the membership factor of pixel $x_j$ to cluster $Y_j$

$d^2(x, y)$ – any distance measure expressing the similarity between a sample data point and the centre of a cluster.

$m$ – the amount of fuzziness of the resulting classification ($1 \leq m < \infty$)

When m=1 $J_m$ generalized to hard partition such as c-means. As m is approaching 1 the algorithm acts more like c-means. As m approaches ∞ the fuzziness factor is more dominant. In case of m=1 formula (2) holds.

$$u_{ij} = \begin{cases} 1, & x_i \in Y_j \\ 0, & otherwise \end{cases} \qquad (2)$$

For $m > 1$, if $x_i \neq c_j$ for all $i$ and $j$, $(U, \bar{c})$ may be locally optimal for $J_m$ only if:

$$u_{ij} = \frac{1}{\sum_{k=1}^{C} \left(\frac{d(x_i, c_j)}{d(x_i, c_k)}\right)^{\frac{2}{m-1}}} \qquad (3)$$

and

$$c_j = \frac{\sum_{i=1}^{N} u_{ij}^m \cdot x_i}{\sum_{i=1}^{N} u_{ij}^m} \qquad (4)$$

Formulas 3 and 4 are used by iterating using simple Picard iteration, by looping back and forth from equation (3) to (4) until there are only small changes of $U$ and $\bar{c}$ between successive iterations[2].

The main disadvantage of the FCM algorithm is susceptibility to noise because of the distance measure $d(x_i, c_j)$ which takes into account only the pixel intensities of the target pixel vs all other pixels in the image and not the surrounding pixels' distance, i.e neighborhood attraction. The result of this limitation is degradation of the algorithms' performance with the addition of noise. As a result the segmentation reaches local minimum and cannot perform the correct noise reduction when noise levels increase above 1%.

Shan[19] proposed an improvement to FCM algorithm called IFCM which introduces neighborhood attraction and neural optimization into the FCM algorithm to overcome the problem of sensitivity to noise. The main addition of the IFCM algorithm is an improved distance measure that takes into account the distance of each pixel evaluated from the target pixel. In FCM the U membership matrix depends heavily on $d(x_i, c_j)$ as shown in formula 3 and $d(x_i, c_j)$ depends only on pixel intensities the membership is highly sensitive to noise. As a result Shan proposed a new distance measure shown in (5):

$$d^2(x_i, c_j) = \|x_i - c_j\|^2 (1 - \lambda H_{ij} - \xi F_{ij}) \quad (5)$$

There are two new parameters introduced due to the balance that need to be achieved between neighborhood attraction and feature attraction. Those two parameters are $\lambda, \xi$ that range between 0-1. In order to find the best segmentation an optimization needs to be executed to find the optimal value of those two parameters. In equation 5 there are also $H_{ij}$ and $F_{ij}$ which represent feature attraction function and neighbourhood attraction function respectively. $H_{ij}$ feature attraction function calculates the feature attraction of pixels in S.

$$H_{ij} = \frac{\sum_{k=1}^{S} u_{kj} g_{ik}}{\sum_{k=1}^{S} g_{ik}} \quad (6)$$

$g_{ik}$ is the intensity difference between the subject pixel $x_i$ and its neighbor pixel $x_k$ as shown in equation 7.

$$g_{ik} = |x_i - x_k| \quad (7)$$

For neighborhood attraction we use equation (8). $q_{ik}^2$ represents the distance between the target pixel k from the neighboring i pixel.

$$F_{ij} = \frac{\sum_{k=1}^{S} u_{kj}^2 q_{ik}^2}{\sum_{k=1}^{S} q_{ik}^2} \quad (8)$$

The relative location between pixel $x_i$ and its neighbourhood pixel $x_k$ is:

$$q_{ik} = (X_{x_i} - X_{x_k})^2 + (Y_{x_i} - Y_{x_k})^2 \quad (9)$$

where $x_i = (X_{x_i}, Y_{x_i}), x_k = (X_{x_k}, Y_{x_k})$ and the neighborhood of $x_i$ is

$$NB_{x_i} = \left\{ x_k \in I : 0 < \left((X_{x_i} - X_{x_k})^2 + (Y_{x_i} - Y_{x_k})^2 < 2^{L-1}\right) \right\} \quad (10)$$

The algorithm is similar to FCM in the iterative manner for which it updates the membership U matrix by using the new distance function as shown in equations (6)-(10).

As a result of the introduction of the new parameters $\lambda, \xi$ the algorithms needs to perform an optimization which includes calculating the cost function in equation (3) each iteration and checking if it was minimized more by choosing different values for those parameters. Shan proposed a neural network optimizer to solve this problem. Some of the disadvantages of IFCM are:

1. Bad choice of initial cluster centers might lead to poor performance.
2. In some cases the runtime is shown to be not optimal due to the complexity of the optimization model.
3. The optimizer might reach local minimum and is no guarantee for optimal values of $\lambda, \xi$ [10][1][15].

To solve these problems different optimization algorithms were used that converge better and may reach global optimum. Next we review the GAIFCM algorithm which uses a genetic optimizer to find best $\lambda, \xi$.

### B. GAIFCM

The GAIFCM[7] uses the GA algorithm and introduces a genome entity for which the IFCM fitness function is being calculated. Each genome can go through the crossover or mutation process. By doing so it modifies the $\lambda$ and $\xi$ parameters which might bring it closer to a global minimum error rate. The algorithm starts by initializing a population of genomes and calculating the fitness function for each. Then by the combination of crossover and mutation each genome is modified and a new population is being generated. For each new population again the fitness function is being calculated until a maximum threshold is reached or the maximum number of generations is reached. Some of the disadvantages of GAIFCM are:

1. The practical runtime of the algorithm suffers from degradation as the size of the image grows.
2. The optimization function still not guaranteed to reach global minimum as noise levels grow.

The algorithm uses standard FCM for the calculation of initial cluster centers to pass to GAIFCM. Since FCM can uses random choice of initialization of cluster centers it can reach local minimum and give sub optimal values to GAIFCM. As a result, the search space that is being explored may not be the optimal.

### C. PSO

Particle Swarm Optimization [15][11] is a nonlinear optimization algorithm introduced by Kennedy and Eberhart in 1995. We chose PSO for this work based on analysis of [9] which shows its improved accuracy and speed over GA. It attempts to simulate social behavior by introducing particles which move in a search space. The particles have velocity and position inside this space. Each particle is able to calculate the cost function being optimized in each iteration of the algorithm. The algorithm holds each particles' best position and the global best position. Each particle can see the neighboring particles and thus chooses to move towards the best particle in its local minima. As a result, the entire swarm moves towards the best positions in the search space. The researchers took the intuition from a swarm of fish or flocking birds after observing their behavior. For example, in the case of a flock of birds each bird can observe the birds next to it and fly in the general direction of the flock. We based our IFCMPSO and 3DPIFCM algorithms on PSO optimization. The algorithm is described below as Algorithm 1.

Algorithm 1: Generic PSO algorithm

| | Input: S – number of particles, d – number of dimensions, f() – optimization function, $\epsilon > 0$ – stop criteria |
| --- | --- |
| | Output: Best particle that minimizes the f() function |
| 1. | For each particle i = 1 …. ,S do |
| 2. | $x_i = U(b_l, b_u)$ – initialize particle's positions uniformly |
| 3. | $p_i = x_i$ – initialize best particle's know position |
| 4. | If $f(p_i) < f(g)$ then |
| 5. | $g = p_i$ – update swarms best position |
| 6. | $v_i = U(-|b_l, b_u|, |b_l, b_u|)$ – initialize particle's velocity |
| 7. | While ch > $\epsilon$ – check if minimal change in position smaller then threshold |
| 8. | For each particle i = 1…..,S do |
| 9. | For each dimension d = 1…, n do |
| 10. | $r_p, r_g = U(0,1)$ – random numbers between 0,1 (Uniform) |
| 11. | $v_{i,d} = \omega v_{i,d} + \varphi_p r_p(p_{i,d} - x_{i,d}) + \varphi_g r_g(g_{i,d} - x_{i,d})$ – update particle velocity |
| 12. | If $f(x_i) < f(p_i)$ then |
| 13. | $p_i = x_i$ – update particle's best position |
| 14. | If $f(p_i) < f(g_i)$ then |
| 15. | $g = p_i$ – update swarms best position |
| 16. | $ch = smallest\ change\ in\ f$ |
| 17. | Return g |

PSO has some advantages to GA in continuous optimization problems in the search space [9]. First, the authors in [9] evaluated PSO vs GA on 8 different sets of problems while measuring quality and computation cost. They reached the conclusion using t-tests that PSO uses less function evaluations than GA while reaching similar quality and thus more efficient. Second, due to the fact that PSO holds two populations for each particle (pbest and current position), this allows for more diversity and exploration then GA.

## III. THE PROPOSED ALGORITHM (3DPIFCM)

### A. Rationale

The development of 3DPIFCM has been done in two stages. In stage one a 2D algorithm was developed which is equivalent to GAIFCM but uses PSO instead of GA. The reason for choosing PSO was for higher speed of convergence [9] and simpler implementation. Also, in the development of IFCMPSO, smarter initialization of the cluster centers was made to avoid local minima.

In stage two a new 3D version of IFCMPSO was developed we call 3DPIFCM. The rationale of moving from 2D to 3D for noise correction was that since medical imaging includes mainly organs that have collocated voxels of same tissue type in 3D space we could utilize those voxels to correct noise during segmentation.

This new algorithm runs iteratively on each voxel in the 2D image much like IFCM but looks at voxels around the target voxel in 3D. When examining each voxel separately in the 3D image we can extract the IFCM features in order to achieve better noise reduction in 2D. In addition, by adding the 3D X order features to the algorithm's attraction function we give higher weight to closer voxels and lower weight to further voxels. We add the 3D features to the attraction equations (6, 8, 9). In figure 2 we show the evolution of the algorithms reviewed and the new algorithms we developed.

Fig 2: evolution of fuzzy clustering algorithms described in this paper. Green boxes are new algorithms presented in this paper.

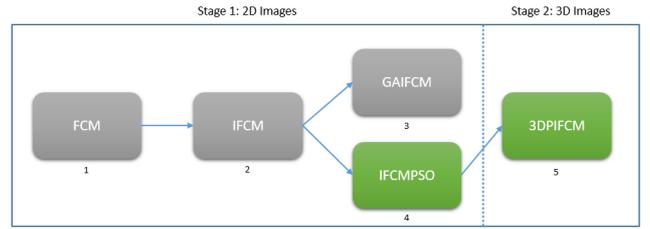

1. Fuzzy c-means algorithm. No noise reduction. Using only feature attraction (pixel colors) to determine clusters.
2. Improved fuzzy c-means. Neural network optimization of neighborhood attraction and feature attraction. New cost function.
3. Genetic algorithm IFCM. Using same cost function as 2 but improving optimization algorithm using GA.
4. PSO based IFCM. Using same cost function as 2 with PSO optimization and change in initialization of cluster centers in modified FCM.
5. 3D version of PSO based IFCM with additional hyper parameters for depth and exponential decay.

### B. Originality and Contribution

Our originality in developing 3DPIFCM comes in two folds:

1. The development of the 2D version named IFCMPSO compares well to GAIFCM but converges faster due to PSO optimization.

2. The 3DPIFCM algorithm is compared to state of the art GAIFCM, IFCMPSO and also FCM for segmentation accuracy in at different levels of noise ranging from 1% to 20%. We can observe that 3DPIFCM outperforms the state-of-the-art by up to 60% at lower level noise down to 13 % in higher levels.

## C. Stage 1: 2D images – IFCMPSO Algorithm Overview

**a. Algorithm Description**

We first describe our modification to the original FCM in algorithm 3:

Algorithm 3: Modified FCM algorithm

| |
|---|
| Input: m – amount of fuzziness, c – number of clusters, $\epsilon$ – stopping threshold, X - image |
| Output: U membership matrix. Each pixel is given a probability to be in each cluster. |
| 1. Fix $m > 1, c \geq 2$ and the stopping criteria $\epsilon > 0$ |
| 2. Initialize $U^{(0)} = [u_{ij}]$ matrix of size $NxC$. |
| 3. **Initialize vector of centers using GMM $C^{(k)} = gmm(c)$** |
| 4. Calculate the vector of centers $C^{(k)} = [c_j]$ using $U^{(k)}$ and equation (4). |
| 5. Update $U^{(k)}$ to $U^{(k+1)}$ using equation (3). |
| 6. If $\|U^{(k+1)} - U^{(k)}\|_\infty < \epsilon$ stop, else repeat steps (4-6). |

Lines 1-2: initialize the U cluster matrix.

**Line 3**: Main change to the original FCM. Instead of initializing vector of cluster centers randomly we run a Gaussian Mixture Model on the image using same number of clusters C. This gives us good estimates and avoids possible local minima that FCM might reach in noisy images.

Lines 4-6: calculate vector centers and update U matrix. Check if stop criteria is reached.

IFCMPSO is described in algorithm 5.

Algorithm 5: IFCMPSO

| |
|---|
| Input: img - a 2D matrix of pixel intensities, c – number of clusters, m - fuzziness, |
|     $\epsilon$ – stop criteria , L – depth level |
| Output: centers, U – membershipMatrix |
| 1. cluster_centers1, U1 = Modified_FCM (img, c, $\epsilon$, m) |
| 2. λ, ξ. cluster_centers2, U2 = PSO (ifcm_step(), cluster_centers1, U1, img, L, m, c) |
| 3. cluster_centers3, U3= IFCM (img , λ, ξ. cluster_centers2, U2,  L, m, c, $\epsilon$) |
|     return cluster_centers3, U3 |

IFCMPSO Algorithm description:

1. A modified version of FCM which uses a Gaussian mixture model [8] for initialization of cluster centers is used instead of a random choice. As a result, local minima is avoided in contrast to optimizations that were produced in earlier experiments.
2. Cluster centers and membership matrix U are used to feed to PSO as initial parameters. In addition, the IFCM_STEP function which performs the optimization according to formulas (1, 3, 4, 5-10) is used. In our description in algorithm 5 λ, ξ are the particles returned from PSO that are optimal to the cost function **ifcm_step()** and image **img**.
3. Once we obtained the optimal parameters λ, ξ for the image a full execution of IFCM algorithm is performed. This finally returns a new membership matrix and cluster centers that are returned by the algorithm.

**b. Assumptions**

1. For M parameter a value of 2 is used. Previous studies show that the optimal value is 1.5 to 2 [7].
2. The value of L=2 is used based on [7] and [19].
3. PSO max iterations is 150. A balanced between performance and quality was necessary and this number was chosen after examining runtime and convergance.
4. Swarm size of 50 was used. Similar to [7] similar hyper parameter for breath of search was used as in GA.

## D. Stage 2: 3D images - 3DPIFCM Algorithm Overview

**a. Algorithm Description**

The 3DPIFCM algorithm uses N[th] order features for segmentation. Those features represent the voxels around the target voxel that is being segmented. The image segmented is still in 2D but voxels in 3D space help to clear the noise. The algorithm iterates over all voxels in the 2D image and in an inner loop examines all voxels in the neighborhood of that center voxel in 3D. It uses feature attraction and neighborhood attraction like IFCM but instead of using a neural network for optimization of λ and ξ it uses PSO. At the start of the algorithm it runs a modified version of FCM that initializes the cluster centers to be in a good position to avoid local minima. In addition, it uses two new parameters h,v as input that control the exponential decay of each neighborhood voxel and the depth of search respectively.

First we define the area around the target voxel as shown in (10) to modified 3D version (10a).

$$NB_{ik} = \left\{ x_k \in I : 0 < \left( (X_{x_i} - X_{x_k})^2 + (Y_{x_i} - Y_{x_k})^2 + (Z_{x_i} - Z_{x_k})^2 \right) < 2^{L-1} \right\} \quad (10a)$$

Figure 6 shows a definition of the features in (10a) for L=1, 2, 3 from left to right.

Fig 6: First, second and third order voxels respectively from left to right. First order has 6 neighboring voxels to center voxel, second has 18 and third has 26. All voxels surrounding the central voxel on the left image are first order. The ones surrounding the central voxel on the central image are second order and voxels surrounding second order are third order.

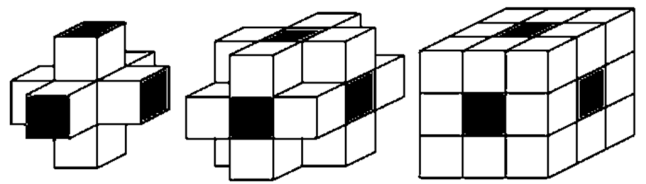

We modify equation (9) to (9a) as follows:

$$q_{ik} = (X_{x_i} - X_{x_k})^2 + (Y_{x_i} - Y_{x_k})^2 + (Z_{x_i} - Z_{x_k})^2 \quad (9a)$$

Equation (9a) looks at each voxel in (10a) and does a squared sum on all axis. This represents the squared difference between the colour of target voxel and a surrounding neighbourhood voxels. We also assign a higher weight to N order voxels and a lower weight to N+1 order voxels. This is done according to w vector which is shown in (6a) and (8a) that are modified versions of (6) and (8) respectively. We introduce a new vector **w** such that the elements of **w** sum to 1. This vector represents the weighted decaying importance of proximity of each voxel according to the originating group.

The size of w vector is v such that $\sum_{r=1}^{V} W_r = 1$. If we assign each element in the vector to be $1/v$ equations (6a) and (8a) will behave exactly like equations (6), (8). The neighboring voxels in (10a) are divided into $v$ sub groups. The v parameter will determine the depth of search for each target voxel. If v > 3 the next set of voxels will be at least 2 voxels away from the target voxel. This may extend the reach of the clustering but may diminish the proximity ingredient. For parameter v a value of 2-5 was experimented with. In part (g) of the results section V we experiment with different values of v. The sub groups are mutually exclusive and $\sum_{r=1}^{V} NB_r \in NB_{x_i}$. The sub groups complement the entire set of neighbouring voxels in (10a).

$$H_{ij} = \sum_{r=1}^{V} W_r \frac{\sum_{k=1}^{S_r} u_{kj} g_{ik}}{\sum_{k=1}^{S_r} g_{ik}} \quad (6a)$$

$$F_{ij} = \sum_{r=1}^{V} W_r \frac{\sum_{k=1}^{S_r} u_{kj}^2 q_{ik}^2}{\sum_{k=1}^{S_r} q_{ik}^2} \quad (8a)$$

We define a decay constant **h**. In order to populate the **w** vector we use an exponential decay equation (11)

$$W_i = \frac{e^{-\frac{i}{h}}}{\sum_{r=1}^{V} e^{-\frac{r}{h}}} \text{ where } i = 1..v \quad (11)$$

Equation (11) is based on half-life exponential decay [20][14]. The formula is meant to lower the importance of each order of voxels as further they are located from the center. For example, if size of vector is v=3 and h=1.1 than the w vector will look like $w = [0.63, 0.25, 0.10]$ thus assigning a higher importance to first order voxels, lower importance to second order and much lower to third. The h parameter is an exponential decay parameter that determines the ingredient of each voxels' contribution to the final distance function according to the group membership.

The higher the h parameter the more weight will be assigned to the furthest voxels and less to the closest. If the h parameter is around 0.5 than 86% of the weight will be to first level voxels closes to the target voxel. On the other hand if the h parameter is very high (for example 100) then w vector will look like $w = [0.33, 0.33, 0.33]$ thus assigning equal importance to each order of voxels from first to n$^{th}$ order. We closely discuss both v and h parameters in section V and experiment with different values to achieve highest performance.

We define the cost function in (12) as the standard FCM cost function $J_m$ with the addition of v and h as parameters. The modified distance function is defined in (6a, 8a, 9a).

$$J_m(U, \bar{c}, h, v, m, \epsilon) = \sum_{i=1}^{N} \sum_{j=1}^{C} u_{ij}^m d^2(x_i, c_j, h, v, m, \epsilon) \quad (12)$$

This cost function already accepts both new parameters h and v and all other standard IFCM parameters m and $\epsilon$. Below we show the flowchart of 3DPIFCM and a more detailed pseudo code for the algorithm. Both are shown in 7a and 7b respectively.

Algorithm 7a: 3DPIFCM algorithm flowchart

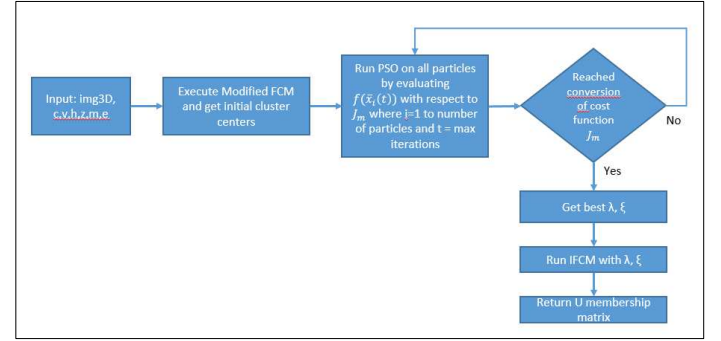

Algorithm 7b: 3DPIFCM Algorithm

Input: img3d - a 3D matrix of pixel intensities, c – number of clusters, v – depth parameter, h – exponential decay, z – which slice in z axis to segment, m – fuzziness, $\epsilon$ – stop criteria
Output: centers, U – membershipMatrix

1. img = img3d[z]
2. cluster_centers1, U1 = Modified_FCM (img, c, $\epsilon$, m)
3. Generate a random swarm of P particles in 2 dimensional space (we use D=2 since there are two optimization parameters).
4. Evaluate fitness of each particle in the swarm $f(\bar{x}_i(t))$ with respect to the cost function $J_m$.
5. If $f(\bar{x}_i(t)) < pbest_i$ then $pbest_i = f(\bar{x}_i(t))$ and $\bar{x}_{pbest_i} = \bar{x}_i(t)$ where $pbest_i$ is the current best fitness achieved by the i-th particle and $\bar{x}_{pbest_i}$ is the corresponding coordinate.
6. 
7. If $f(\bar{x}_i(t)) < lbest_i$ than $lbest_i = f(\bar{x}_i(t))$, where $lbest_i$ is the best fitness over the topological neighbors.
8. Change the velocity $v_i$ of each particle: $\bar{v}_i(t) = \bar{v}_i(t-1) + p_1(\bar{x}_{pbest_i} - \bar{x}_i(t)) + p_2(\bar{x}_{lbest_i} - \bar{x}_i(t))$ $p_1$ and
9. $p_2$ are random constants between 0 and 1.
10. Fly each particle to its new position $\bar{x}_i(t) + \bar{v}_i(t)$
11. Go to step 4 until convergence (i.e small changes to J cost function).

cluster_centers2, U2, λ, ξ = $\bar{x}_{pbest_i}$ variables.

|    |    |
|----|----|
|    | cluster_centers, U= IFCM (img , λ, ξ. cluster_centers2, U2, v, m, c, ϵ) |
| 12 | Return cluster_centers, U |

Algorithm 7 description:

1. Assign the z slice to a new variable
2. Run the modified FCM with Gaussian mixture model instead of random initialization of centers.
3. Generate random particles to evaluate parameters λ, ξ.
4. Steps 4-9 are running the PSO algorithm and executing the step function of 3DPIFCM at each step. The step function includes evaluating equation 1 with modified formulas 6a, 8a, 9a.
10. Take the best particle in the swarm after the swarm finished executing and get values λ, ξ.cluster centers and membership matrix.
11. Execute standard IFCM with correct λ, ξ. parameters and correct membership matrix and cluster centers.
12. Return the U membership matrix and cluster centers.

b. **Hyper parameter values**

We make certain assumptions in all implementations of variants of FCM. The hyper parameters we use for 3DPIFCM are shown in table 9. The choice of hyper parameters builds upon the works of [2][6][7][19]. For the new hyper parameters we introduce in this work (shown in table 10) we explore their effect on performance in sections (g) and (e).

Table 9: hyper parameters values

| Hyper Parameter | Value | Description | Algorithms |
|---|---|---|---|
| M | 2 | Fuzziness | All |
| E | 0.01 | Stop criteria | All |
| L | 2 | 2D Neighborhood level | GAIFCM, FCM, IFCMPSO |
| Max iterations | 150 | Another stop criteria in case we don't reach epsilon | All |
| PSO swarm size | 50 | Number of particles in the swarm | 3DPIFCM, IFCMPSO |
| omega | 0.5 | Velocity scaling factor | 3DPIFCM, IFCMPSO |
| phip | 0.5 | Scaling factor to search away from the particle's best known position | 3DPIFCM, IFCMPSO |
| phig | 0.5 | Scaling factor to search away from the swarm's best known position | 3DPIFCM, IFCMPSO |
| Pso_max_iter | 20 | The maximum number of iterations for the swarm to search | 3DPIFCM, IFCMPSO |
| minstep | 1e-8 | The minimum step size of swarm's best position before the search | 3DPIFCM, IFCMPSO |
| minfunc | 1e-8 | The minimum change of swarm's best objective value before the search | 3DPIFCM, IFCMPSO |

In addition to the static hyper parameters we research the dynamic hyper parameters that are new to 3DPIFCM. We show their ranges in table 10 and examine their results in respect to the segmentation performance in section 4.

Table 10: dynamic hyper parameters ranges for 3DPIFCM

| Hyper Parameter | Name | Value range | Description |
|---|---|---|---|
| h | Exponential decay | 0.01 - 100 | Determines the ingredient of each voxels' contribution to the final distance function according to the group membership. |
| v | Depth | 2 – 5 | Determines the depth of search for each target voxel. |

c. **Assumptions**

The main premise of our new algorithm is that image comes in 3D format. This is most common in medical imaging domain such as MRI and CT scan where there are 3 planes of view. Another assumption is that the images come in greyscale colors. This is also most common format in medical imaging. Although our algorithm should be able to handle RGB colors, this option was not tested during this work.

Some assumptions on hyper parameters are:

1. Same assumptions as in C.b hold here for 3DPIFCM.
2. The new parameter h values of 0.01 to 100 were chosen. Those values represent experiments that were performed in results section V.
3. In the new Depth parameter values of 2-5 were chosen. Those values represent experiments that were performed in results section V.

In addition, the algorithm was test image sizes ranging from 32x32 to 854x854. Standard medical image sizes come in 512x512 pixels in MRI which are well within those bounds. Although there is no theoretical or computational limit to the size of the images we assume standard medical sizes.

E. *Data set definitions*

Two different datasets were used to test the accuracy of the segmentation. The first dataset is a synthetic dataset that includes 4 squares of different colors in grayscale one inside the other. The volume is the same size in pixels as Brainweb[4] which is 181x217x181 voxels. This is the standard size of the Brainweb[4] volume that is published for research and we copied it for comparison. Gaussian granular noise was added to the synthetic data ranging from 16% to 27%. The high noise ratio is because the data is homogeneous and all algorithms perform very well under lower noise levels for this set. The synthetic data is shown in figure 11.

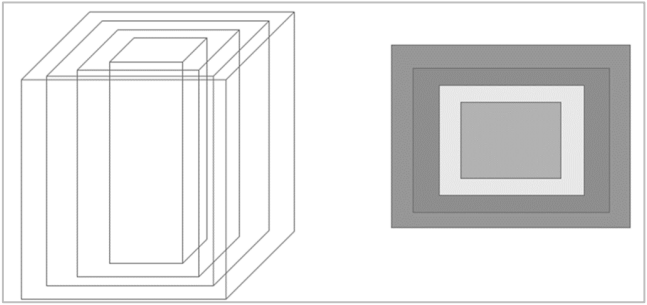

Fig 11: on the left is a 3D representation of the synthetic data of size 181x217x181 voxels. Each cube will have a difference color. On the right is a top view of the cube.

The second data set is Brainweb[4] T1 simulation of an adult brain. several volumes of varying amounts of noise were created ranging from 1% to 20%. Both Gaussian and Poisson noise in same levels were tested since Gaussian is more prevalent for MRI scans and Poisson for CT scans. All noise levels were homogeneous in our tests. In addition, we examined the variability of noise and statistical significance of the algorithm's performance. Figure 12 shows a sample of Brainweb[4] data.

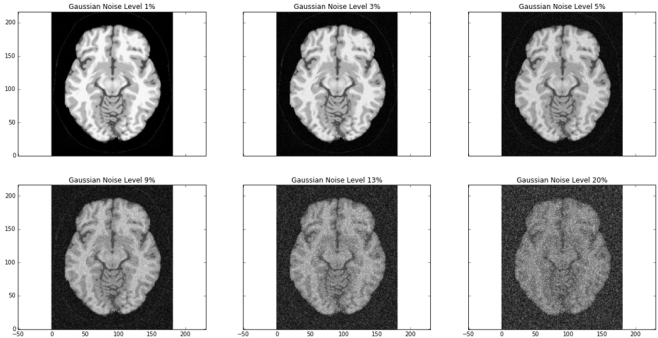

Fig 12: Brainweb[4] data with Gaussian noise level from 1% to 20%.

As can be seen in the Brainweb[4] data in higher noise levels the image becomes unclear and tissues are indistinguishable.

### F. Development toolchain

The implementation of the algorithms in all cases was done in python 2.7 with numpy for numeric computations to speedup performance. In addition, we used the numba[18] package to JIT compile the functions that make up the most costly operations such as the step functions in FCM, IFCM, 3DPIFCM and GAIFCM. The step functions go through the entire image pixel by pixel and perform the calculations of the clusters and depicted in formulas 1-11 inclusive.

The JIT compilation step makes the implementation nearly identical to a native C/C++ variant. We wanted to make the algorithms run as fast as possible but still use a high level scientific language. Although in this paper we are not analyzing the speed of the implementations in terms of raw second performance it was still necessary to write optimized code to increase number of experiments and test for hyper parameters. The research stack also included using nibabel[5] for reading medical image file types such as Nifti and matlab matrices in MINC[16] format. In order to visualize the results the Jupyter Notebook stack was used.

### IV. RUNTIME ANALYSIS

We analyze the asymptotic performance of 3DPIFCM algorithm together with IFCMPSO. Analysis of GAIFCM. IFCM, FCM exist in [7]. We compare this data to the new algorithm. First we examine the performance of FCM as stated in [7][2]. FCM runs an iterative process which has max number of iterations that are defined by a parameter.

For each iteration there is an evaluation of the entire image. We assume image size N pixels and cluster size C. Each membership calculation of FCM takes $\theta(C^2 \cdot N)$. As shown in [7][2] the overall asymptotic performance of FCM is therefore $O(\text{maxIterations} \cdot C^2 \cdot N)$.

IFCM modifies the distance function to account for noise. A new parameter S is added as a neighbourhood parameter that accounts for distance measure from target pixel being evaluated in each iteration to all pixels in S. the overall performance of IFCM is shown to be

$$\max\{O(\text{maxIterations} \cdot S \cdot C \cdot N), O(\text{maxIterations} \cdot C^2 \cdot N)\}.$$

Since IFCM uses FCM as an initial segmentation to get initial cluster centres we take the maximum of either the running time of FCM or the evaluation of IFCM with the S parameter.

GAIFCM as described in [7] uses a population and generation parameters. It runs an iterative process by which all the population is affected and each genome is running a full step function of IFCM which evaluates to $O(S \cdot C^2 \cdot N)$.
If we assume S a small constant, eventually Frizon et al shows that the effective asymptotic performance of the algorithm is

$$\Theta(\text{generationsCount} \cdot \text{populationSize} \cdot C^2 \cdot N)$$

In order to evaluate 3DPIFCM we first evaluate IFCMPSO since both algorithms work with the PSO optimization function.

#### 1. IFCMPSO analysis

Particle Swarm Optimization works by first selecting the size of the swarm which we denote SW. The swarm consists of particles that have a position and velocity in the search space of the swarm. Each particle has a dimension which consists of the number of parameters the algorithm is trying to optimize.

1. First the algorithm runs FCM which is $\theta(C^2 \cdot N)$
2. The main loop is iterating each particle in the swarm for each dimension. For each iteration the objective function is evaluated which has a cost of $\Theta(C^2 \cdot N)$.
3. During the test of the objective function there is an update to both velocity and position of the swarm according to

$$v_{i,d} = \omega v_{i,d} + \varphi_p r_p (p_{i,d} - x_{i,d}) + \varphi_g r_g (g_{i,d} - x_{i,d}).$$

The position is updated by evaluating the objective function and getting the result of segmentation.
4. This process runs max number of iterations until minimal change detected or max iterations is reached.

The 4 steps shown above and fully in table 1 define the asymptotic performance to be $\Theta(SW \cdot d \cdot maxIterations \cdot C^2 \cdot N)$. As in our case the d parameter is not significant (1 or 2) the formula is reduced to $\Theta(SW \cdot maxIterations \cdot C^2 \cdot N)$ Asymptotically this implies that GA and PSO have similar performance. Nevertheless, research shows in [9] that PSO outperforms GA by at least factor of 2 due to faster convergence.

## 2. 3DPIFCM analysis

Our new 3DPIFCM algorithm also uses PSO as an optimizer but adds a new dimension to the proximity and feature equations. This measure is the S parameter which indicates how many voxels to examine in each iteration of the modified IFCM step function. Since we are using a depth of 2 – 5 we can calculate the number of voxels involved by each iteration of the step function.

As depicted in algorithm 7 we analyse 3DPIFCM.
Lines 1-3: generating a random swarm of particles and running FCM with GMM as cluster centres. This action is a standard FCM evaluation which is $O(maxIterations \cdot C^2 \cdot N)$
Lines 4-9: we evaluate the objective function which is evaluated by each particle in the swarm SW number of times with d dimensions. The IFCM objective function was $\Theta(C^2 \cdot N)$. In the case of 3DPIFCM the cost is $\Theta(S \cdot C^2 \cdot N)$. The swarm is evaluated maxIterations or until convergence where the stop criteria is met.

Therefore, the final evaluation of the algorithms is $\Theta(SW \cdot S \cdot d \cdot maxIterations \cdot C^2 \cdot N)$
Since we define early on that S and d are negligible the evaluation becomes $O(SW \cdot maxIterations \cdot C^2 \cdot N)$ which is similar to GAIFCM asymptotically.
Nevertheless as stated in [9] and section 4.2.1 the observed runtime performance is faster by a factor of 2.

## V. RESULTS

### 1. Evaluation parameters

In order to perform a quantitative evaluation of the results three definitions were used in as in the original Shen paper[19]. The evaluation parameters are chosen since we are doing clustering and seek to find if a particular pixel or voxel is in the correct cluster. Table 13 shows the segmentation evaluation parameters.

Table 13: Evaluation parameters

| Name of measurement | Formula | Description | Comments |
|---|---|---|---|
| Under Segmentation | $UnS = \dfrac{N_{fp}}{N_n}$ | The percentage of positive false segmentation for a cluster. | $N_{fp}$ – The number of pixels that do not belong to a cluster and are segmented into that cluster (false positives). $N_n$ – The total number of pixels that do not belong to that cluster. |
| Over Segmentation | $OS = \dfrac{N_{fn}}{N_p}$ | The percentage of negative false segmentation for a cluster. | $N_{fn}$ – The number of pixels that belong to a cluster and are not segmented into that cluster (false negatives). $N_p$ – The total number of pixels that belong to that cluster. |
| Incorrect Segmentation | $IncS = \dfrac{UnS + OS}{N}$ | The total percentage of false segmentation. | |

All our comparisons and evaluations of algorithms are done by the IncS measure. This measure combines both Under Segmentation and Over Segmentation. It is averaged by the total number of pixels/voxels being segmented.

### 2. Evaluation procedure

The results section shows a comparison of all algorithms in for different noise types and different data types. The purpose of this section is to evaluate the new algorithms in comparison to known implementations. Following evaluations were performed:

a. We will first define the evaluation parameters for the segmentation quantitatively as defined in the original Shen paper [19].
b. We will run our 3DPIFCM algorithm against synthetic data as shown in figure 11. We will evaluate using only homogeneous Gaussian noise for this part. The reason for this is that the synthetic data is Uniform and not resembling real data. We perform different noise types such as Gaussian and Poisson in the Brainweb data instead which simulates MRI scans although Poisson noise is more relevant for CT scans.
c. We evaluate the algorithm on Brainweb data using both Gaussian and Poisson noise types in different levels.
d. We perform comparative analysis of 3DPIFCM qualitatively against GAIFCM, IFCMPSO, FCM using only incorrect segmentation measure (incS) since it provides a good measure of overall model accuracy.
e. We summarize the experiments made on h,v hyper parameters.
f. We analyze the experiments on H parameter which is the exponential decay of voxel's contribution to final clustering and discuss results.
g. We analyze the experiments on V parameter which is the depth of search between target voxel and search voxel and discuss results.
h. We present a showcase with visual results comparing 3DPIFCM to the best performing 2D version GAIFCM. We show the segmentation of brainweb data and compare example executions in different noise levels.
i. We perform a runtime analysis of all algorithms discussed in this paper.

## 3. Running 3DPIFCM on synthetic data

### 3.1 Purpose

The main goal of running the new algorithm on synthetic data was to test the ability of the algorithm to generalize to different types of data. We ran multiple tests on synthetic volumes presented in section III (E).

### 3.2 The experiment

We executed the algorithm in different noise levels with varied amounts from 1% to 27% noise. In this experiment only Gaussian noise type was used. An experiment containing an additional noise type was conducted on the synthetic brain data to simulate real world conditions.

### 3.3 Results

We witnessed that below 16% there was no use to show results since the algorithm performed perfect segmentation. Moreover, when testing other FCM variants including FCM we saw the same results below 16% noise. As can be shown in figure 14 on synthetic data under 16% noise there were almost no segmentation errors. The data in the synthetic volume is highly homogeneous and uniform between voxels in the same cluster. Since each cube sites within another outer cube it's easier for the algorithm to use neighborhood attraction of the voxels to determine the right segment. Despite the good results shown in the synthetic volume we expected much more realistic results when looking into Brainweb data which simulates real world brain.

Fig 14: 3DPIFCM on synthetic volume, 10-27% Gaussian noise

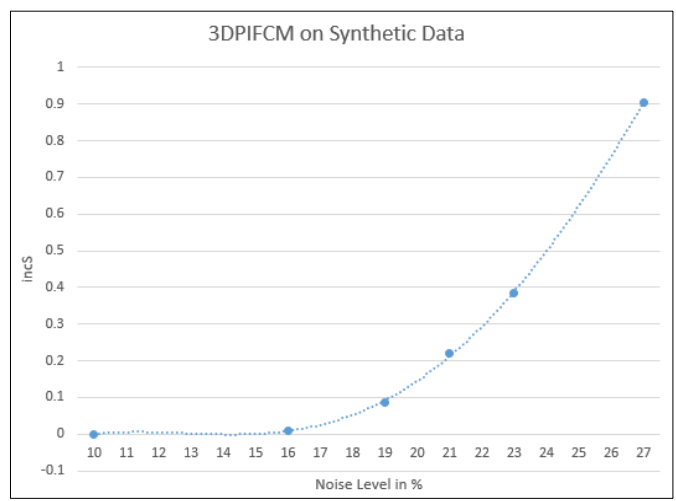

### 3.4 Analysis

We can see from the results in figure 14 that in highly correlated images the algorithm which specialized in noisy images performs well. When 16% noise is reached we can see errors start to occur in segmentation accuracy. This is highly indicative of the stability of the algorithm in different conditions. Although using synthetic generated data is not indicative of real world conditions in medical imaging it gives a good indication of the generality of the algorithm to changing textures and image types. Also, by using highly uniform data we provide a simple test case to tackle before moving on to simulated brain.

The analysis shows that 3DPIFCM behaves very well on synthetic data and compares well to the original FCM and IFCM variants. There is a steady increase in error rates as noise levels grow beyond 16% showing stable behavior.

## 4. Running 3DPIFCM on Brainweb data

### 4.1 Purpose

The experiment was done with simulated brain volume that gives us ground truth. Since clustering was performed we still require the labels of the pixels/voxels to give us the accuracy of the algorithm. Different noise levels presented by Brainweb ranging from 1% to 20% were used.

The purpose in using both Gaussian and Poisson noise types for those experiments was to accommodate real world conditions which include different image modalities such as CT and MRI. It is known fact that CT scans mainly exhibit poisson noise[3] while MRI can have Gaussian. In addition, we wanted to measure the effect of different noise types and noise levels on the algorithms' performance in order to assess its ability to generalize to other parts of body and other type of modalities.

### 4.2 The experiment

We executed the algorithm on Brainweb T1 slice 60 Axial view. We chose this slice as a good middle slice in axial view when the dimension of the entire brainweb volume was 181x217x181. We used hops of 2% in noise increase from 1% to 9% and additional two executions for 13% and 20% noise. This is so that large noise levels will be examined. Nevertheless, larger noise levels of 13% and 20% are very uncommon as shown in original Shen paper[19].

### 4.3 Results

Results show that there is almost a linear relationship between noise levels and error rates when testing against Brainweb data. The two noise types are almost identical for executions of up to 9% where we see a small increase in error rate from 9% onwards for Gaussian noise.

Fig 15: 3DPIFCM on Brainweb T1 volume Slice 60, 1-20% noise (Gaussian and Poisson noise).

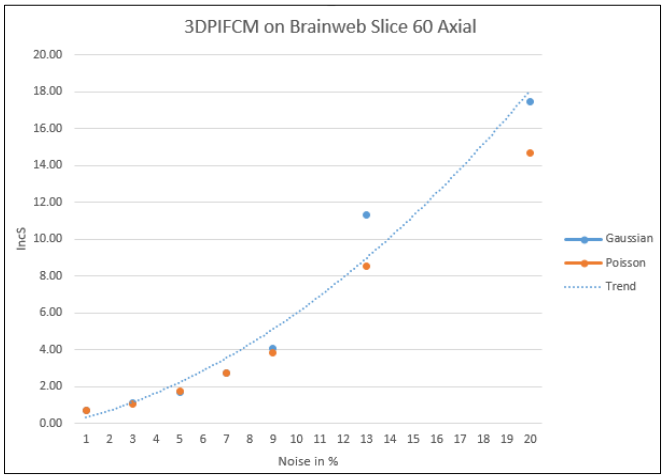

### 4.4 Analysis

We can see that the algorithm behaves well in lower noise levels and increases error rates linearly as noise increases. As shown in figure 15, between 13% and 20% noise levels it's very difficult to distinguish between the real voxel information and noise artefacts. Since we focus on real equipment noise (1%-9%) this is acceptable and our goal is to compare this performance to other FCM variants in 2D including FCM itself. As noise level increase from 9% onwards there is clearly better segmentation results for Poisson noise by an increasing margin.

The results indicate that both Poisson and Gaussian noise types are similar in segmentation performance as shown on Brainweb data. Above 9% noise there is a slight advantage to Poisson but since the signal to noise ratio is very low in those levels it is non consequential. Also, a regression line can be seen to fit to the incS as noise levels increase. This indicates that there is a linear relationship between segmentation performance and noise levels.

## 5. *Comparative analysis of running 3DPIFCM against FCM, IFCMPSO, GAIFCM*

### 5.1 Purpose

We evaluate 3DPIFCM on Brainweb data as shown in section c. We compare the performance of the algorithm using IncS to 3 different algorithms running on 2D images of the same slice. Our algorithm is the only one that is utilizing 3D information as well as the 2D slice. Our main purpose is to compare qualitatively the accuracy of segmentation with varied amounts of Gaussian and Poisson noise levels. We perform the comparison between each algorithm to 3DPIFCM with formula 13.

$$IncS^{A-3DPIFCM} = \frac{IncS^A - IncS^{3DPIFCM}}{IncS^A} \cdot 100 \qquad (13)$$

In formula (13) A is denoted as the the algorithm for comparison. This formula indicates the relative percentage improvement or decrease of 3DPIFCM performance compared to algorithm A. This comparison is done against FCM, IFCMPSO and GAIFCM

### 5.2 The experiment

As before we evaluate on Gaussian and Poisson noise for all algorithms. We use 1,3,5,7,9,13,20 percent noise levels for all algorithms and compare by IncS error. Figures 16 and 17 show our results on executions of Gaussian and Poisson noise levels respectively. In figure 18 we show a similar experiment that was done on the synthetic data shown in III (E).

### 5.3 Results

Results show that when comparing to FCM both in Gaussian and Poisson noise we see a dramatic improvement in lower noise levels of 1% up to 9%. The improvement according to (13) is up to 65% peeking in 9% noise and diminishing in 13% noise and higher. When comparing to GAIFCM and IFCMPSO the improvement ranges between 5-20%. Figure 16 shows the comparison on Gaussian noise and figure 17 on Poisson. The negative numbers in figures 16-18 indicate competing algorithms outperform 3DPIFCM by a percentage shown. In figure 18 we show the same experiment and comparison on simulated data shown in section V.E. We can see an average difference in 31% and 22% against GAIFCM and IFCMPSO respectively across noise levels. Also, as noise grows from 16% to 27% in synthetic data we see a continuous decline in improvement suggesting that noise to signal ration is rapidly diminishing for those noise levels.

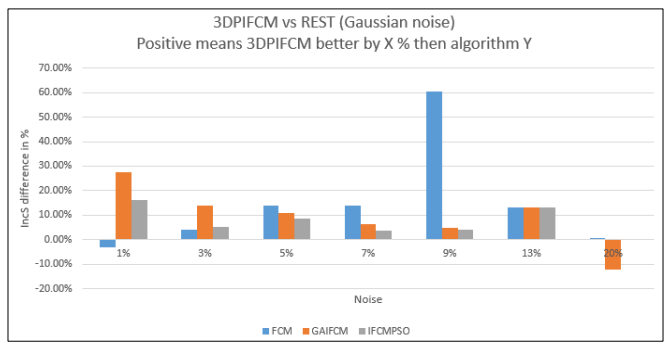

Fig 16: Comparison of 3DPIFCM vs FCM, IFCMPSO and GAIFCM at 1-20% Gaussian noise using (13).Using Brainweb T1 Volume slice 60. At noise level 20% GAIFCM outperforms 3DPIFCM by 11%.

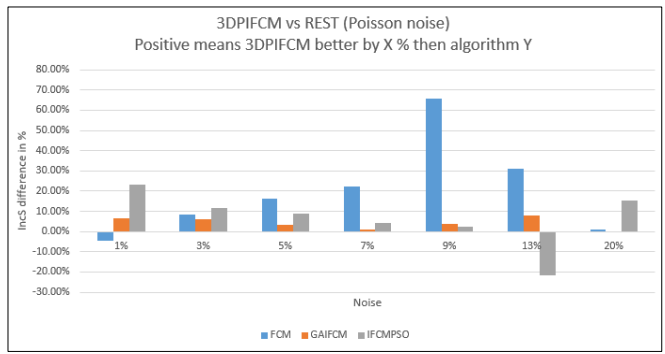

Fig 17: Comparison of 3DPIFCM vs FCM, IFCMPSO and GAIFCM at 1-20% Poisson noise using (13). Using Brainweb T1 Volume slice 60.

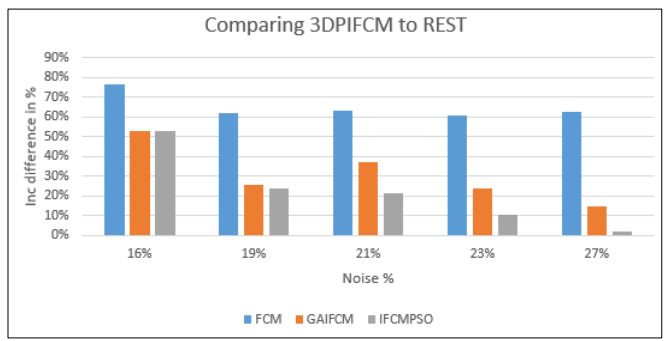

Fig 18: Comparison of 3DPIFCM vs FCM, IFCMPSO and GAIFCM at 16-27%, synthetic data with Gaussian noise using (13).

## 5.4 Analysis

When we analyze the performance of 3DPIFCM against GAIFCM we see an interesting trend. In lower noise levels of 1-13% there is an average improvement of 12% to 3DPIFCM for Gaussian noise. In Poisson noise the situation is a little less dramatic and we see an average of 5% improvement for same levels. In 20% noise levels there is an advantage to GAIFCM of 15% and 12% for Gaussian and Poisson respectively.

Nevertheless, even FCM only performs at a similar level to 3DPIFCM at 20% noise. This suggests that the 3D nature of the algorithm which collects voxels in 3 dimensions per iteration abstracts its performance at very high noise levels. Also, as can be seen in both [19] and in our paper section V.E, in 20% the image is almost indistinguishable from noise.

In synthetic data the results show similar characteristics. We perform the comparison on 16%-27% and see a decline in improvement as noise levels grow to 27% and beyond

## VI. EVALUATING NEW DYNAMIC HYPER PARAMETERS

### 1. Purpose

As described in table 10 there are two new hyper parameters in 3DPIFCM. The effect of those parameters on the overall accuracy of the algorithm was researched. Each parameter was analyzed separately. To do this we modified other parameters during the executions and took the average or minimum execution for each parameter value depending on the experiment performed.

### 2. Hyper parameter H

#### 2.1 The experiment

The 3DPIFCM algorithm was run on Simulated brain data T1 Axial view on slice 60. We selected this slice as it is a central slice of the brain and representative for segmentation. The H parameter was changed, and all other values remained the same for the sake of this experiment. Values ranging from 0.01 to 100 were chosen. After a look on the effect of this parameter we can see that a low value such as 0.01 will give a very high weight to the first order of voxels and a much lower weight to next levels. The effect will be that closer voxels will affect the segmentation of the target voxel much more then further ones. In the same manner if we use an H value of 100 than the weigh is distributed evenly between far voxels and close voxels thus having similar effect on the target voxel segmentation.

In our experiments for each value of H we executed a series of runs of the algorithm with different noise levels ranging from 1%-20%. In figure 19 we can see the execution of the algorithm on noise levels 1%-9%. We omit the 13%-20% because we there was a very high standard deviation on the results of each execution in different H levels and because as mentioned in previous sections at those noise levels the algorithm reduces accuracy to near FCM level.

## 2.2 Results

Fig 19: Execution of 3DPIFCM with different values of H ranging between 0.01 and 100. The experiment ran on Brainweb T1 Volume in slice 60 Axial view with Gaussian noise.

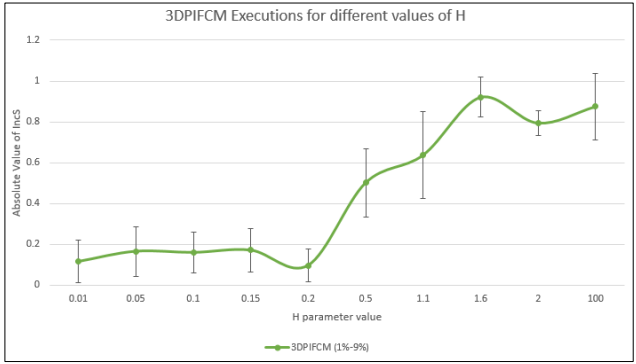

## 2.3 Analysis

There is a saddle point when a value of 0.2 is reached. This suggests that there is a local optimum for this value in Brainweb data for different noise levels. Also, above 0.2 there is a steady increase in error rates regardless of the amount of noise suggesting that if we distribute the weights between far and near voxels more evenly it will reach a point that clustering performance will be degraded.

Since there is a local optimum for h parameter in this data type, it can be assumed that for similar data types of other modalities we could also find local optimums. As a result, different values of h can be preconfigured for different modalities and body types in images.

## 3. Hyper parameter V – 3D Depth

### 3.1 The experiment

In this experiment a noise levels increased between 1%-20% and the best performing V value was taken for each noise level which minimized incS. If the H parameter described the weight of X order voxels from target voxel, V represents the depth of the search per voxel in each iteration of the algorithm. In this experiment values of V ranged between 2 and 5 while all other parameters remained the same. In our earlier tests we saw that a larger number than 5 will significantly increase computation time and not improve performance.

If we calculate the number of voxels that need to be examined per iteration for the v value the numbers are as follows, for v=2 there are 18 voxels in the search space, v=3 there are 26 voxels, v=4 there are 92 voxels and finally for v=5 there are 116 voxels. This is an exponential series and computation time is affected significantly if we go beyond 5. Figure 20 shows the effect of the V parameter for different noise levels ranging from 1% to 20%. All experiments ran on Brainweb T1 Volume slice 60.

### 3.2 Results

Fig 20: Execution of 3DPIFCM on Brainweb T1 slice 60 Gaussian noise for different noise levels taking the minimum for each V value.

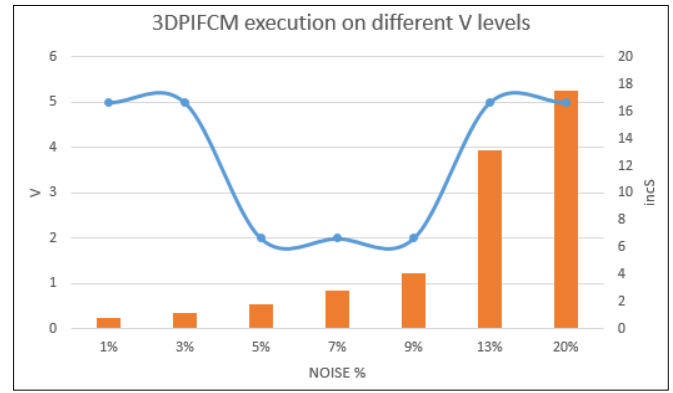

### 3.3 Analysis

We can see that for lower noise levels the best performance is gained by using a higher v. For middle noise between %5-%9 a low value is more performant. In high noise levels of %13-%20 we can witness an additional rise in this parameter. This suggests that high noise levels are not sensitive to the depth parameter since the noisy voxels block the information from reaching the central voxel being evaluated. As a result we could formulate the pre selection of this value depending on known noise level as default.

The results of this experiment indicate that there is an optimal value for the v parameter per noise level and per modality type. It can be concluded that a preselection step can be performed according to noise and image types and default values can be chosen before executing 3DPIFCM on a new data set.

## VII. CASE STUDY

### a. Purpose

One of our goals was to test 3DPIFCM qualitatively against the other algorithms with varied noise levels. To perform such a test we segmented an adult brain slice in Brainweb[4] with all algorithms and took the WM (white matter) segmentation as a test case. The reasoning behind this approach is to view the results more clearly with a naked eye and compare which pixels were misclassified in each algorithm. Figure 21 shows a comparison of 3DPIFCM against GAIFCM for WM. Again, for sake of brevity we remove the 1% noise comparison for lower bound and over 9% noise for upper bound. Results in those bounds are very difficult to visualize and provide no added value to our analysis.

b. **Results**

Fig 21: comparison of 3DPIFCM and GAIFCM with different noise levels ranging from 3% to 9% with Gaussian noise.

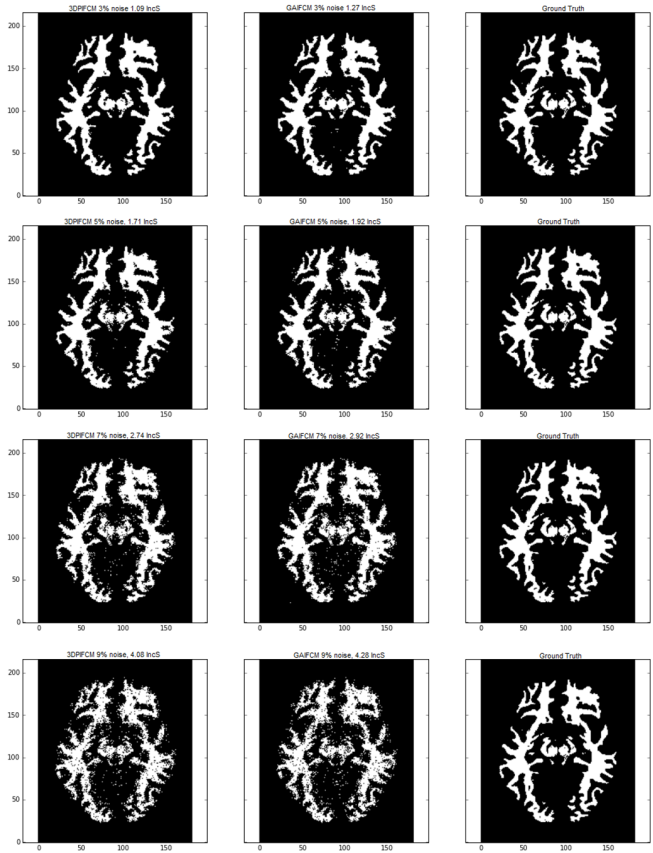

c. **Analysis**

We can see in the results that edge cases where the clusters are close to each other 3DPIFCM prevails because of the higher noise reduction capability as a result of 3D corrections. From 13% onwards the quality of image degrades to a level that it's almost indistinguishable between noise artifacts and real pixels. For this reason in figure 21 we show only noise levels of 3%-9%. We omitted 1% noise since results were almost indistinguishable and not visible enough. Also, above 9% as we showed in d. there is no clear advantage to 3DPIFCM mostly because of low noise to signal ratio.

In conclusion, all comparisons in the case study show that the mean of error difference between 3DPIFCM and GAIFCM is 0.19 in incS with standard deviation of 0.012. This means that there is a constant gap of 0.2% absolute error rate reduction in advantage to 3DPIFCM compared to GA. In addition, it can be seen that the error is evenly distributed in the images. This indicates that the performance increase is not due to local advantage but as a result of either the optimization algorithm or the additional 3D features that counter noise.

VIII. LIMITATIONS

The limitations of our study were:

1. Usage of only simulated brain data for real world example but not actual brain MRI. This was because it was critical to measure accuracy in comparison to other algorithms based on accurate labeling, which came with simulated Brainweb data.
2. Comparison was made with similar clustering algorithms using different optimization techniques to account for noise. We didn't test accuracy against a completely different segmentation paradigm like region growing or supervised deep learning.
3. Using python with scientific package and JIT compilation might not be the most optimal language to test for speed when comparing similar algorithms.

Overall our study represents a general approach to clustering given noisy artifacts in medical images. The limitations presented above could all be mitigated by data enhancement, change of hardware and software.

IX. CONCLUSIONS

In this paper we introduced a new segmentation algorithm that is based on IFCM and particle swarm optimization. The purpose of this algorithm was to gain accurate segmentation in 3D images that contain noise due to intensity inhomogeneity or a bias field such as MRI and CT scans. The algorithm is unsupervised and requires no training data. As a result, it can be a good fit for clustering images where there is a small number of labeled examples available. The algorithm uses 3D features in order to counter the noisy slices produced by the equipment. To evaluate the performance of this algorithm we tested on both synthetic and Brainweb data.

We showed that the algorithm outperforms state of the art genetic variants such as GAIFCM and IFCMPSO by a margin of 5-50% improvement for synthetic data and 1-28% improvement for simulated brain data using the standard incorrect segmentation which is the percentage of false segmented pixels in the image. We compare Gaussian and Poisson noise functions separately on all the datasets to gain higher confidence on segmentation ability and noise reduction. We showed that our algorithm utilizes the 3D nature of the image hence gaining more information about the surrounding voxels during each iteration and as a result performs with higher quality. Also, we showed that the algorithm works best for medium noise levels 3-10% which are most common as shown in Brainweb. The results show that in very low noise of 0-1% there is no advantage over FCM or other genetic variants because the attraction features are not performing noise reduction and in fact reduce accuracy.

In addition, we observed that in very high noise levels of 20% and above there is no significant difference between standard FCM and the noise reduction variants. We suspect that this is because of very low signal to noise ratio which prevents noise reduction to be effective. In addition to introducing 3DPIFCM we also evaluate two new hyper parameters which control the depth of search from target voxel and the exponential decay of a contribution of each voxel to the overall clustering. We evaluated those parameters in different noise levels and show that there can be optimum values depending on noise levels and image type. This suggests that the algorithm can be preconfigured with default values in the future according to image types and noise types.

We also present a case study of qualitative results of WM comparison between GAIFCM and 3DPIFCM. We see that in areas in between clusters GAIFCM can misclassify. This suggests that our 3D version which can look at top and bottom slices in the z index can account of the noise artifacts better the 2D variants and therefore is more accurate around the edges and in between the clusters.

## X. Further Research

In this research we evaluated 3DPIFCM against other genetic segmentation algorithms such as GAIFCM and IFCMPSO that reduce noise from 2D images. We used Simulated Brain MRI images which normally come in 3D image formats such as Nifti or MINC. We would like to evaluate the algorithm on different sets of organs with ground truth both in CT scans and MRI scans to gain better understanding of the generalization ability of the algorithm in different conditions. Additional qualitative evaluation would be to check segmentation results by doctors and qualified medical staff to further establish credibility and assess the quality of the algorithm.

In future research we would like to explore ways to speed up the computation of this algorithm using various techniques. First, we would like to explore the usage of parallelization of both data and computational resources on a single 2D image or a whole 3D volume. Second, the usage of better runtime environments such as compiled languages and code optimization techniques can achieve improvements.